\documentclass{article}

\usepackage{PRIMEarxiv}

\usepackage[utf8]{inputenc} 
\usepackage[T1]{fontenc}    
\usepackage{hyperref}       
\usepackage{url}            
\usepackage{booktabs}       
\usepackage{amsfonts}       
\usepackage{nicefrac}       
\usepackage{microtype}      
\usepackage{lipsum}
\usepackage{fancyhdr}       
\usepackage{graphicx}       
\usepackage{amsmath}
\graphicspath{{media/}}     

\title{CORONA-Fields: Leveraging Foundation Models for Classification of Solar Wind Phenomena}

\author{%
 Daniela Martin\thanks{These authors contributed equally to this work.}\\
 University of Delaware \\
 \texttt{dmartinv@udel.edu}
 \And
 Jinsu Hong\footnotemark[1] \\
 Georgia State University \\
 \texttt{jhong36@gsu.edu} \\
 \And
 Connor O'Brien \\
 Boston University \\
 \texttt{obrienco@bu.edu} \\
 \And
 Valmir P. Moraes Filho \\
 Catholic University of America \\
 \texttt{moraesfilho@cua.edu} \\
 \And
 Jasmine R. Kobayashi\\
 Southwest Research Institute \\
 \texttt{jasmine.kobayashi@swri.org} \\
 \And
 Evangelia Samara \\
 NASA Goddard Space Flight Center \\
 \texttt{evangelia.samara@nasa.gov} \\
 \And
 Joseph Gallego \\
 Drexel University \\
 \texttt{jg3959@drexel.edu} \\
}

\begin{document}
\maketitle

\begin{abstract}
Space weather at Earth, driven by the solar activity, poses growing risks to satellites around our planet as well as to critical ground-based technological infrastructure. Major space weather contributors are the solar wind and coronal mass ejections whose variable density, speed, temperature, and magnetic field make the automated classification of those structures challenging.
In this work, we adapt a foundation model for solar physics, originally trained on Solar Dynamics Observatory imagery, to create embeddings suitable for solar wind structure analysis. These embeddings are concatenated with the spacecraft position and solar magnetic connectivity encoded using Fourier features  which generates a neural field-based model. The full deep learning architecture is fine-tuned bridging the gap between remote sensing and in situ observations. Labels are derived from Parker Solar Probe measurements, forming a downstream classification task that maps plasma properties to solar wind structures. Although overall classification performance is modest, likely due to coarse labeling, class imbalance, and limited transferability of the pretrained model, this study demonstrates the feasibility of leveraging foundation model embeddings for in situ solar wind tasks. As a first proof-of-concept, it lays the groundwork for future improvements toward more reliable space weather predictions. The code and configuration files used in this study are publicly available to support reproducibility.
\end{abstract}

\begin{figure*}[t]
\includegraphics[width=\linewidth]{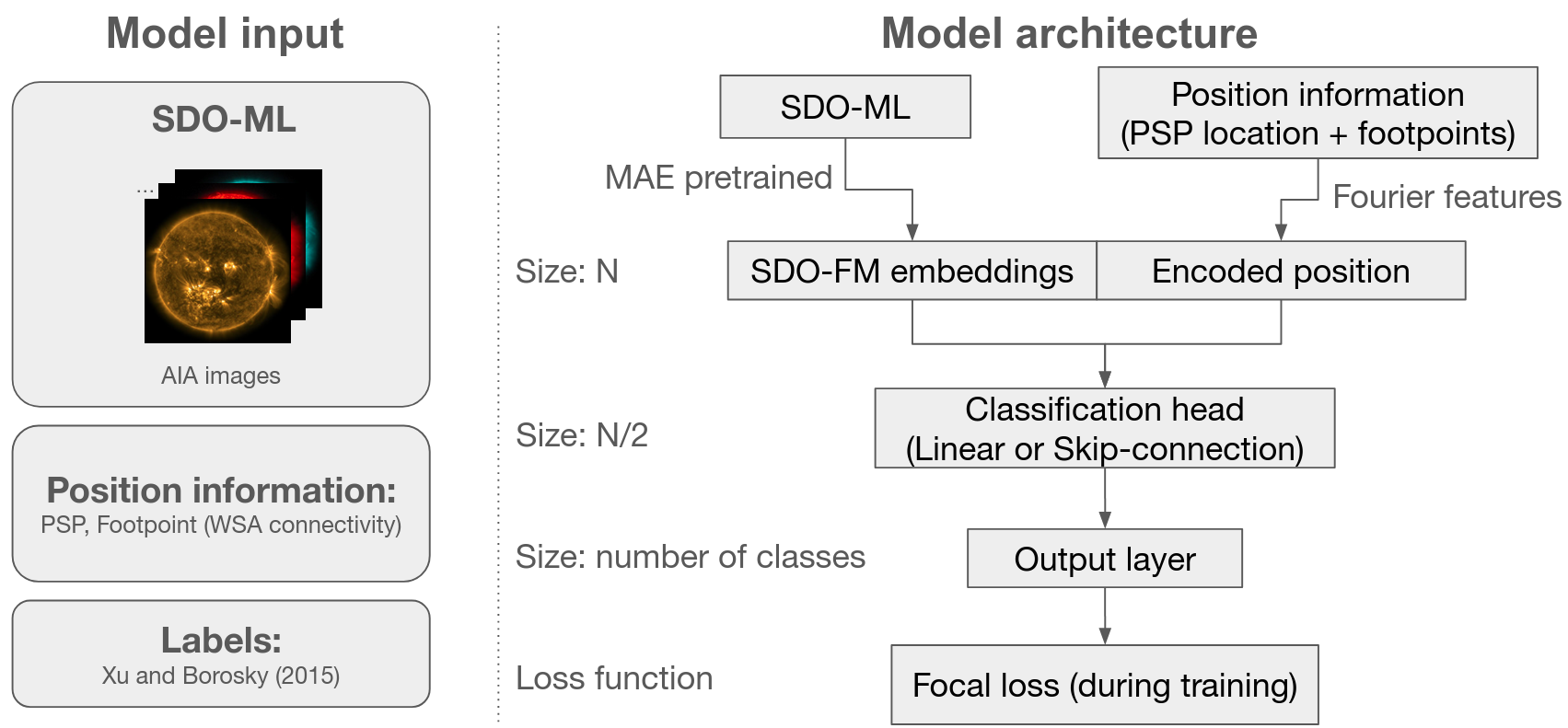}
\caption{Architecture of the proposed model. The framework combines two input streams: (i) solar images processed by a pretrained MAE backbone (SDO-FM) to obtain an embedding layer, and (ii) positional information from the Parker Solar Probe and its magnetic footpoints, encoded through Fourier features. The embeddings and encoded positions are concatenated and passed through the classification head (either a simple linear head or a skip-connection variant), producing logits over the available solar wind classes. During training, predictions are compared with ground-truth labels using the focal loss function.}
\label{fig:model-arch}
\end{figure*}

\section{Introduction}
The solar wind is the continuous outflow of charged particles from the Sun into the interplanetary space \cite{parker_dynamics_1958}. It is composed of a variety of structures that differ in density, speed, temperature, magnetic field, and other properties depending on their source region on the Sun. Particularly, the fast component of the solar wind, otherwise known as high speed streams, that originates from coronal holes on the Sun, can cause medium- to large-scale geomagnetic storms \cite{richardson2001, echer2013}. Such storms can disrupt satellites and pose radiations risks to astronauts. Within the solar wind, more extreme and transient phenomena frequently propagate such as coronal mass ejections (CMEs). CMEs are large expulsions of plasma and magnetic field from the Sun’s corona \cite{forbes2000, webb2012} that produce variations in the solar wind and, in turn, cause variations in the Earth’s magnetic field as well \cite{richardson_sources_2000}. CMEs can cause large-scale geomagnetic storms which cannot only severely damage satellites \cite{baruah_loss_2024} and pose even higher risks than high speed streams to astronauts' life, but they can also disrupt terrestrial power grids \cite{hamrin_space_2023} leading to significant economic consequences \cite{eastwood_economic_2017}. Reliably forecasting both solar wind and CMEs remains highly challenging as our understanding of their structure and evolution is still incomplete \cite{owens_rate_2022}.

Classifying solar wind structures (from now on when referring to solar wind we mean solar wind and CMEs together), is central to space weather research as solar wind structures originating from different sources (e.g., coronal holes, streamer belts, active regions, etc.) have distinct plasma and magnetic properties that determine their geoeffectiveness \cite{borovsky_spatial_2018}. Being able to distinguish and predict these structures is key to reliably forecast geomagnetic activity and mitigate risks related to satellites, power grids, and astronauts. Several studies have attempted to predict solar wind properties directly from solar disk images. For example, \cite{upendran_solar_2020} used EUV images to forecast wind speed, while \cite{lin_prediction_2023} employed magnetograms. More recently, \cite{wang_forecasting_2025} predicted solar-wind-driven geomagnetic activity from EUV images. While these studies show promise, they typically rely on physics-based preprocessing of solar images before using them in neural networks. This step introduces uncertainty from model choices and may discard valuable information contained in the raw data.

Recent progress in self-supervised learning has enabled the training of foundation models, namely, large neural networks pretrained on raw, unlabeled data, learn general-purpose representations useful for many tasks \cite{darcet2023vision}. Notable examples include SimCLR \cite{chen2020simple}, DINO \cite{caron2021emerging}, and MAE \cite{he2022masked} foundation models. Specifically for MAE, it learns compact latent embeddings by reconstructing masked portions of the input. These embeddings can then be reused in downstream applications. MAE has already demonstrated strong transfer performance in fields like Earth observation \cite{allen2023fewshot, gallego2023exploring}, motivating their use in solar physics.


In this work, our goal is to design a deep learning framework to connect solar imagery with in situ solar wind observations (see overall architecture in Figure~\ref{fig:model-arch}). We leverage the SDO Foundation Model (SDO-FM) \cite{walsh2024foundation}, a pretrained masked autoencoder (MAE) trained on solar images as a foundation model backbone for the downstream task of solar wind classification. To enhance representational power, we incorporate a neural field-based head architecture that encodes positional information in a high-frequency space, exploring both simple linear heads and more expressive variants based on ResNet \cite{he2016deep} and NeRF \cite{mildenhall2021nerf} skip connections. Recent work has highlighted the importance of enriching models with positional information. Incorporating this idea into our setup enables solar wind classification to take into account the geometric context of the heliosphere, beyond what is available in the image embeddings alone. Solar wind structures are labeled using in situ plasma measurements mapped to the \cite{xu_new_2015} scheme. Although overall classification performance is modest, the model appears to capture meaningful structure in the solar wind data, suggesting that foundation model embeddings can encode relevant physical information. Taken together, our results demonstrate that combining foundation models with in situ labeling schemes offers a first step toward effectively linking solar imagery with heliospheric plasma data, being a nontrivial step toward unifying solar physics and space weather forecasting. 


\section{Observational datasets}

To close the gap in our knowledge of the solar wind, the Parker Solar Probe (PSP) \cite{fox_solar_2016, szabo_flying_2018} was launched in August 2018. One of the main objectives of this mission is the measurement of the solar wind properties at distances closer to the Sun than any other mission has achieved before (as close as $\approx9$ Solar Radii $(R_{Sun})$ or $\approx0.04$ Astronomical Units (AU)). That makes it the only spacecraft that has recorded in situ properties of the solar wind at an extended radial range within the inner heliosphere ($\approx0.04AU$ to $\approx1.0AU$; see \cite{raouafi_parker_2023} for a summary of findings from PSP's initial mission). PSP carries four instrument suites—FIELDS, SWEAP, WISPR, and IS$\odot$IS—designed to measure different aspects of the solar wind. In this study, we use the FIELDS instrument \cite{bale_fields_2016}, which measures the in situ magnetic field components 
and the SWEAP instrument \cite{kasper_solar_2016}, which measures the properties of the solar wind plasma.


In order to try to connect PSP in situ measurements to their solar source regions, we use remote-sensing images from the Solar Dynamics Observatory (SDO) \cite{Pesnell2012}. SDO was launched in 2010 and observes the solar atmosphere with a suite of instruments, including the Atmospheric Imaging Assembly (AIA) \cite{lemen_atmospheric_2012} and the Helioseismic and Magnetic Imager (HMI) \cite{schou_design_2012}. AIA acquires 10-channel images of the solar disk (the full visible surface of the Sun as seen from Earth) at two ultraviolet (UV), seven extreme ultraviolet (EUV), and one visible wavelength \cite{Lemen2011}, while HMI observes the solar disk to derive global photospheric velocity, intensity, and magnetic field measurements \cite{Hoeksema2014}. 




Although AIA and HMI provide high-resolution 4096~×~4096 images, such data is not yet well-suited for direct application in multi-modal learning with machine learning or deep learning algorithms \cite{jarolim2025deep}. To address this, \cite{galvez2019machine} introduced the SDOML dataset, a clean and machine-learning-oriented version of SDO observations downscaled to 512~×~512, resulting in the reduction of computational demands. The SDOML dataset was designed to provide solar physicists and data scientists with a standardized, machine-learning-ready collection of SDO observations. To achieve this, raw Level-1 data from the three instruments on board SDO were first corrected to remove known instrumental artifacts such as flat-field irregularities, exposure variations, and bad pixels. Spatial and temporal downsampling was then applied to reduce the very high resolution of the original data to a more computationally manageable scale, while still preserving structures of interest for scientific and machine learning tasks. All image sequences were temporally aligned and co-registered across wavelengths to ensure that simultaneous multi-channel observations correspond to the same solar disk region and time frame, which is critical given the varying cadences of SDO instruments. Metadata, including helioprojective coordinates, solar ephemeris, and instrumental context, were retained in standardized formats, allowing downstream tasks to connect observations across physical domains. The resulting dataset therefore eliminates the need for extensive preprocessing at the user level, offering “analysis-ready” images where instrumental and observational inconsistencies have already been corrected. By leveraging SDOML, we can directly use compact embeddings from foundation models without repeating ad-hoc preprocessing pipelines \cite{walsh2024foundation}, which reduces uncertainty and ensures reproducibility across experiments. We use images from SDOML provided at a cadence of 12 minutes, matching the downsampled format described in \cite{galvez2019machine}, sufficient to capture large-scale coronal structures relevant to solar wind origin while keeping the dataset computationally manageable.

\section{Models} \label{sec:model}
\subsection{Backbone: SDO-FM}
Several approaches have been proposed to build foundation models with SDO data. The SDO-FM model \cite{walsh2024foundation} utilizes SDOML and leverages Masked Autoencoder (MAE) \cite{he2022masked} and the Nouveau Variational Autoencoder (NVAE) \cite{vahdat2020nvae} to construct the backbone. We build upon the SDO-FM \cite{walsh2024foundation}, a large-scale masked autoencoder (MAE) trained on the SDOML dataset from 2010–2023 using TPU clusters. This model serves as a pretrained backbone that encodes rich spatiotemporal representations of solar structures. During pretraining, random patches of the input images are masked, and the encoder learns to produce a compact latent representation that allows the decoder to reconstruct the missing regions. This self-supervised approach enables MAE to capture meaningful spatial and structural features of the solar corona \cite{he2022masked, walsh2024foundation}. The output embedding layer from the encoder is used with a custom head neural network to do the downstream task of solar wind classification. 

The SDO-FM model used in this work was previously trained with AIA data from the SDOML dataset. Given the nature of its training, it is expected that it will primarily capture coronal intensity features rather than magnetic topological ones, which may limit its ability to identify potential magnetic drivers of the solar wind. The specific model was selected as it was already pretrained, allowing us to explore downstream solar wind classification while acknowledging its limitations.

\subsection{Architecture of the Classification Head}
To enhance the model's representational power, we incorporate neural fields, positional information from the heliospheric PSP's latitude and longitude and the photospheric footpoint of the magnetic field line that PSP is connected to based on Wang-Sheeley-Arge (WSA) \cite{arge_improvement_2000} predictions, through sinusoidal (Fourier) features into a high-dimensional spectral space \cite{rahimi2007random, mildenhall2021nerf}. The mapping of a coordinate $x$ is defined in Equation~\ref{pos-encode}, where $L$ is the number of frequency bands. This encoding captures both low- and high-frequency variations, providing richer information about the spacecraft’s position and its connection to the solar surface.

\begin{equation}
\gamma(x) = \Big[ \sin(2^0 \pi x), \cos(2^0 \pi x), \dots,  
            \sin(2^{L-1} \pi x), \cos(2^{L-1} \pi x) \Big]
    \label{pos-encode}
\end{equation}

The MAE embeddings and positional encodings are concatenated to form a unified representation, which serves as the input to the classification head. Two head variants were explored.

\paragraph{Linear head.}  
The linear head consists of three fully connected layers with ReLU activations and dropout. The second and third layers use half the number of neurons of the preceding layer, respectively.

\paragraph{Skip-connection head.}  
The skip-connection head follows the strategy of NeRF \cite{mildenhall2021nerf}, where intermediate layers receive both the hidden activations and the original input through skip connections every $k$ layers. This allows the model to preserve fine-grained positional information across deeper transformations and enables richer interactions between pretrained embeddings and positional encodings. All the head's layers have the same number of neurons.

Both architectures produce a final layer with four logits corresponding to the classes. Together, these two variants allow us to assess the trade-off between simplicity and expressive capacity in integrating positional information into the solar wind classification pipeline.

\subsection{Training strategy} \label{train-strategy}

Our training follows a two-stage transfer learning and fine-tuning strategy:

\begin{enumerate}
    \item \textbf{Transfer learning stage:} we initialize the backbone with the pretrained SDO-FM weights and freeze all its layers, training only the newly added classification head (and positional embedding layers) to adapt the representations to the solar wind classification task.

    \item \textbf{Fine-tuning stage:} after convergence of the head, we unfreeze the entire network, including the MAE transformer backbone, allowing joint optimization of all parameters to refine the representations for in situ plasma classification.
\end{enumerate}

Training is performed with the Adam optimizer \cite{kingma2014adam} and a plateau learning rate scheduler. To handle class imbalance, we use the focal loss \cite{lin2017focal}, defined as $\text{FL}(p_t) = -\alpha_t (1 - p_t)^\gamma \log(p_t)$, where $p_t$ is the predicted probability for the true class, $\alpha_t$ balances class weights, and $\gamma$ emphasizes hard-to-classify samples \cite{lin2017focal}. This loss encourages the model to learn discriminative features even for underrepresented solar wind categories. As shown in Figure \ref{fig:model-arch}, the concatenated features are passed through the classification head, where the focal loss is applied during both training stages.

\section{Experiment}
In this section, we describe the preprocessing steps and evaluation protocols used to assess our model.

\subsection{Preprocessing of PSP datasets}
PSP measurements from the FIELDS and SWEAP instruments were first resampled to a common one-minute cadence via binning and averaging. SWEAP has a native cadence of approximately 25 seconds, while FIELDS measures at $\approx$ 3 ms; thus, each one-minute bin contains roughly 20,000 FIELDS and 2-3 SWEAP measurements. Minutes with missing data were linearly interpolated and flagged per instrument to allow later filtering if necessary. Additional parameters were calculated from PSP measurements to enrich the dataset. As mentioned earlier, the magnetic footpoint of PSP on the solar surface and the corresponding solar wind travel time were obtained using the WSA model \cite{arge_improvement_2000}. These footpoints allow each measurement to be linked to the region of the Sun from which the measured solar wind parcel originated from. WSA was used for ease of calculation and as a baseline for image selection.

Solar wind structure labels were derived using a four-class segmentation scheme based on solar wind plasma conditions \cite{xu_new_2015}. This classification partitions solar wind into fast plasma originating from coronal holes \cite{sheeley_coronal_1976}, slower wind plasma from the streamer belt region \cite{crooker_global_2012}, dense and cold sector reversal plasma forming the heliospheric current sheet \cite{susino_physical_2008}, and ejecta which are linked to CMEs and magnetic clouds \cite{richardson_sources_2000, zhao_global_2009}. Labels are based on proton temperature, proton specific entropy, and Alfvén speed, providing a notional link to the solar origin of each wind parcel. These four structural classes serve as the labels for the classification algorithm developed in this study. The structure of the algorithm is easily adaptable to other classification schemas; this one was chosen for its computational simplicity and notional link to the approximate area the solar wind originated from. It should be acknowledged, however, that this scheme presents challenges and has big uncertainties because of the fact that (1) the connectivity of the ejectas' footpoints back to the Sun based on the WSA model are not reliable since the WSA model predicts only solar wind and no transient structures, and (2) the fact that the streamer belt and sector reversal categories often display overlapping properties resulting in some inconsistencies and overlapping in earlier classifications \cite{camporeale2017classification}.

\subsection{Connectivity of PSP measurements with SDOML images}
In order to connect PSP measurements with an SDOML/AIA image, we assumed that WSA produces a good estimate of the solar wind at each PSP location. Then, for each such PSP location, we traced back to the photosphere the magnetic field line that the spacecraft was connected to. This is how each PSP measurement was paired with a SDOML/AIA image, producing a dataset of image-PSP position-footpoint-label. A single SDOML image can be associated with multiple PSP observations due to higher PSP dataset cadence when compared it against SDOML dataset. However, this does not introduce label noise into the dataset, since each pairing corresponds to a distinct heliospheric position and magnetic footpoint. In practice, this provides the model with complementary views of different source regions from the same solar image enhancing the diversity of spatial information available during training. 

The training set includes data from April to December between 2019-2023. The validation set uses data from January to March between 2019-2022. The test set consists of data from January to March of 2023. Overall, the training, validation, and test sets comprise approximately 92\%, 6\%, and 2\% of the dataset, respectively, ensuring temporal separation to avoid potential leakage due to solar rotation. Despite representing about 2\% of the total dataset, the test split remains statistically meaningful given the overall dataset size of nearly one million samples, each composed of 10-channel AIA imagery. This corresponds to over 13{,}000 test instances, which provides a sufficiently large sample to evaluate model generalization with low variance in performance metrics. The choice of using approximately three months of data for testing was deliberate: by reserving a contiguous temporal segment, the evaluation avoids data leakage across solar rotations and ensures that the test period is unseen both spatially and temporally. This design prioritizes independence and realism over uniform proportional splitting, which is critical in temporally correlated heliophysics datasets.

\begin{figure*}[htbp]
    \centering
    \includegraphics[width=0.98\linewidth]{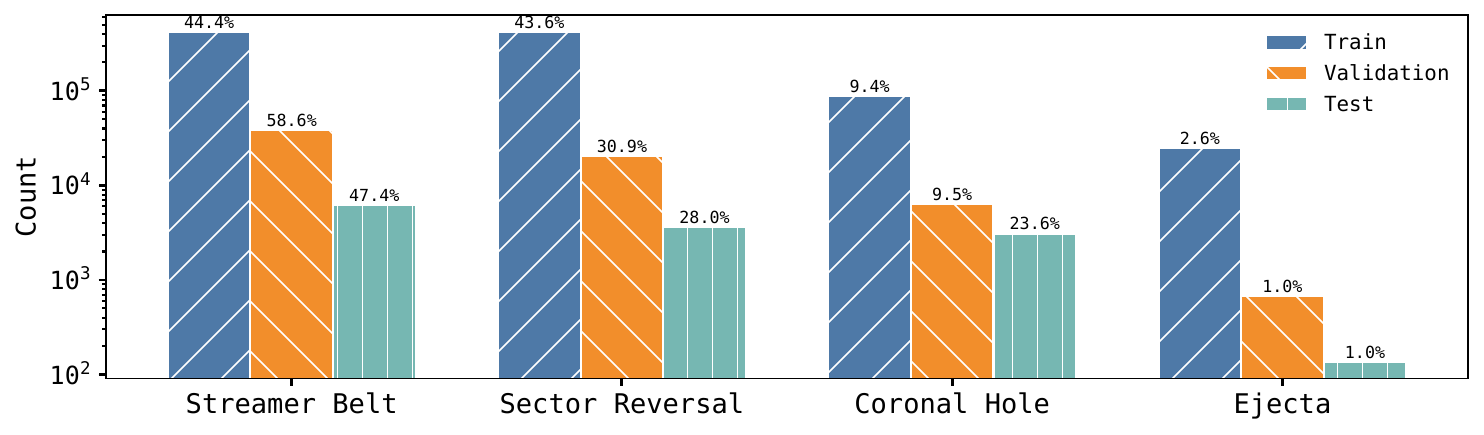}
    \caption{Distribution of each solar wind class across each dataset.}
    \label{fig:class_dist}
\end{figure*}

Table~\ref{tab:dataset-splits} summarizes the dataset partitions and class distributions. Figure~\ref{fig:class_dist} shows the inherently imbalanced nature of the data: streamer belt wind and sector reversal dominate, while coronal hole and ejecta are underrepresented, motivating the use of focal loss during training.

\begin{table*}[htbp]
\centering
\begin{tabular}{lrrrrr}
\hline
Set & Total & Streamer Belt & Sector Reversal & Coronal Hole & Ejecta \\
\hline
Train & 953{,}821 & 415{,}870 & 423{,}960 & 89{,}206 & 24{,}785 \\
Validation & 66{,}245 & 38{,}799 & 20{,}444 & 6{,}319 & 683 \\
Test & 13{,}148 & 6{,}235 & 3{,}675 & 3{,}102 & 136 \\
\hline
\end{tabular}
\caption{Dataset partition and class distribution. Each sample corresponds to a 10-channel AIA image and its associated plasma parameters.}
\label{tab:dataset-splits}
\end{table*}

\subsection{Training setup and evaluation}
To evaluate model performance in the imbalanced multi-class classification task, we report accuracy, precision, and F1-score, all computed per class and averaged with a macro strategy. To mitigate class imbalance, we tuned the $\alpha$ and $\gamma$ parameters of the focal loss. Models were trained for 50 epochs with a batch size of 32, and hyperparameters (including learning rate, weight decay, and scheduler) were selected based on focal loss validation. The full hyperparameter search space is summarized in Table~\ref{tab:hyperparams}. All experiments were implemented in PyTorch (FP32) and executed on Google Cloud Platform using four c2-standard-8 VM instances (8 vCPUs, 32 GB RAM) with 10 TB SSD storage and four NVIDIA A100 GPUs. The full codebase supporting this work is available at \url{https://github.com/spaceml-org/CORONA-FIELDS}.

\begin{table*}[h]
\centering
{%
\begin{tabular}{ll}
\hline
Parameter  & Search Space                                                         \\ \hline
Head Type           & Linear layers, Linear layers with skip-connection                              \\
Optimizer           & Adam                                                                           \\
Hidden Layer Size   & 64, 128, 256, 512, 1024                                                        \\
Learning Rate       & $1 \times 10^{-5}$, $1 \times 10^{-6}$, $1 \times 10^{-7}$, $1 \times 10^{-8}$ \\
Weight Decay        & $3 \times 10^{-4}$, $1 \times 10^{-4}$, $1 \times 10^{-3}$                     \\
Scheduler           & Cosine Annealing, Reduce on Plateau                                            \\
Loss Function       & Cross-Entropy, Focal Loss                                                      \\
Focal Loss $\alpha$ & {[}0.45, 0.30, 0.15, 0.10{]}, {[}0.45, 0.35, 0.10, 0.10{]}                     \\
Focal Loss $\gamma$ & 2, 3                                                                           \\
Sampling Strategy   & No modification, Under-sampling                                                \\ \hline
\end{tabular}%
}
\caption{Hyperparameter search space for the solar wind classification task.}
\label{tab:hyperparams}
\end{table*}

\subsection{Results}

\begin{figure}[htbp]
    \centering
    \includegraphics[width=0.7\linewidth]{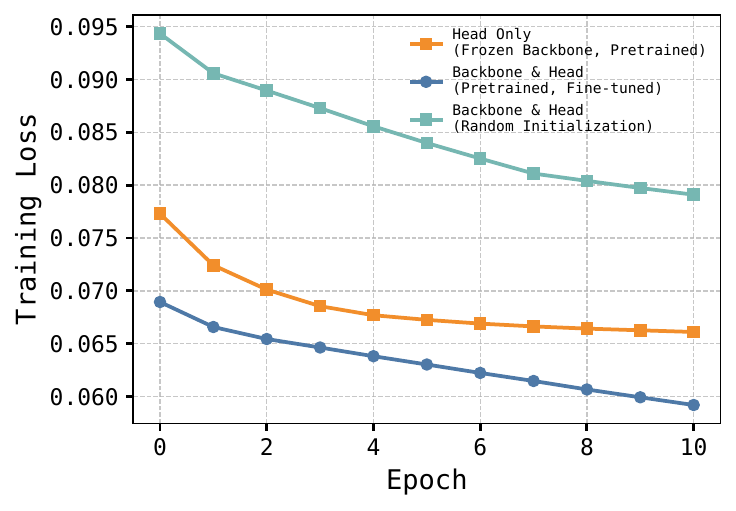}
    \caption{Training loss curves for a linear head with skip-connection. Three model configurations are compared: (1) head only, with a frozen, pretrained backbone, (2) backbone \& head with a pretrained backbone fine-tuned during training, and (3) backbone \& head with random initialization.}
    \label{fig:training-loss}
\end{figure}

Figure \ref{fig:training-loss} compares training loss across three configurations. The model with random initialization of the MAE backbone exhibits consistently higher loss across all epochs compared to the other two configurations. This outcome is expected, as leveraging a pretrained backbone provides prior knowledge about the image domain and avoids retraining it from scratch on potentially massive datasets, thereby facilitating faster and more effective optimization. Among the models with lower loss, the configuration with a fine-tuned pretrained backbone achieves the best performance, indicating that adapting pretrained features with new task-specific information leads to more effective learning. Overall, the comparison highlights the value of pretraining for stable and efficient optimization. Training was stopped early based on validation loss, as part of an early-stopping criterion to prevent overfitting once performance on the validation set plateaued.

\begin{figure*}[htbp]
    \centering
    \includegraphics[width=0.95\linewidth]{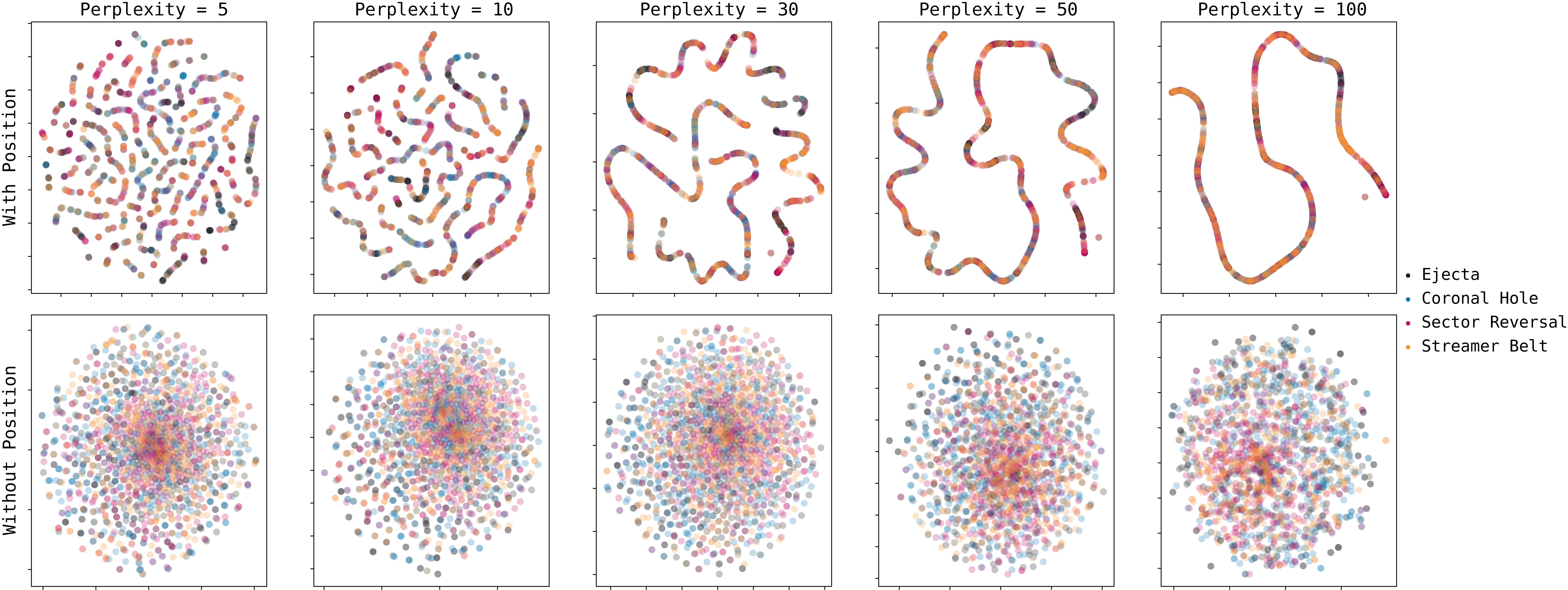}
    \caption{t-SNE on a randomly selected balanced set of embeddings produced by SDO-FM with PSP's position included (top row) and without PSP's position included (bottom row) at five different perplexities. With positional information included, the embeddings are projected into a lower-dimensional space, as evidenced by the strip-like structure in the t-SNE visualization. }
    \label{fig:tsne}
\end{figure*}

In addition to evaluating training loss, we also analyzed how incorporating positional information affects the structure of the embedding space. After adding positional encoding, the embeddings are projected into a lower-dimensional space, as visualized in the t-SNE plots of embeddings with and without positional encoding (Figure \ref{fig:tsne}). While the different types of solar wind structures are not fully separable in either case, adding positional information reveals interesting patterns and a clearer organization of the embedding space, reflecting a dimensionality reduction induced by the positional encoding.

\begin{figure*}[ht]
    \centering
    \includegraphics[width=0.9\linewidth]{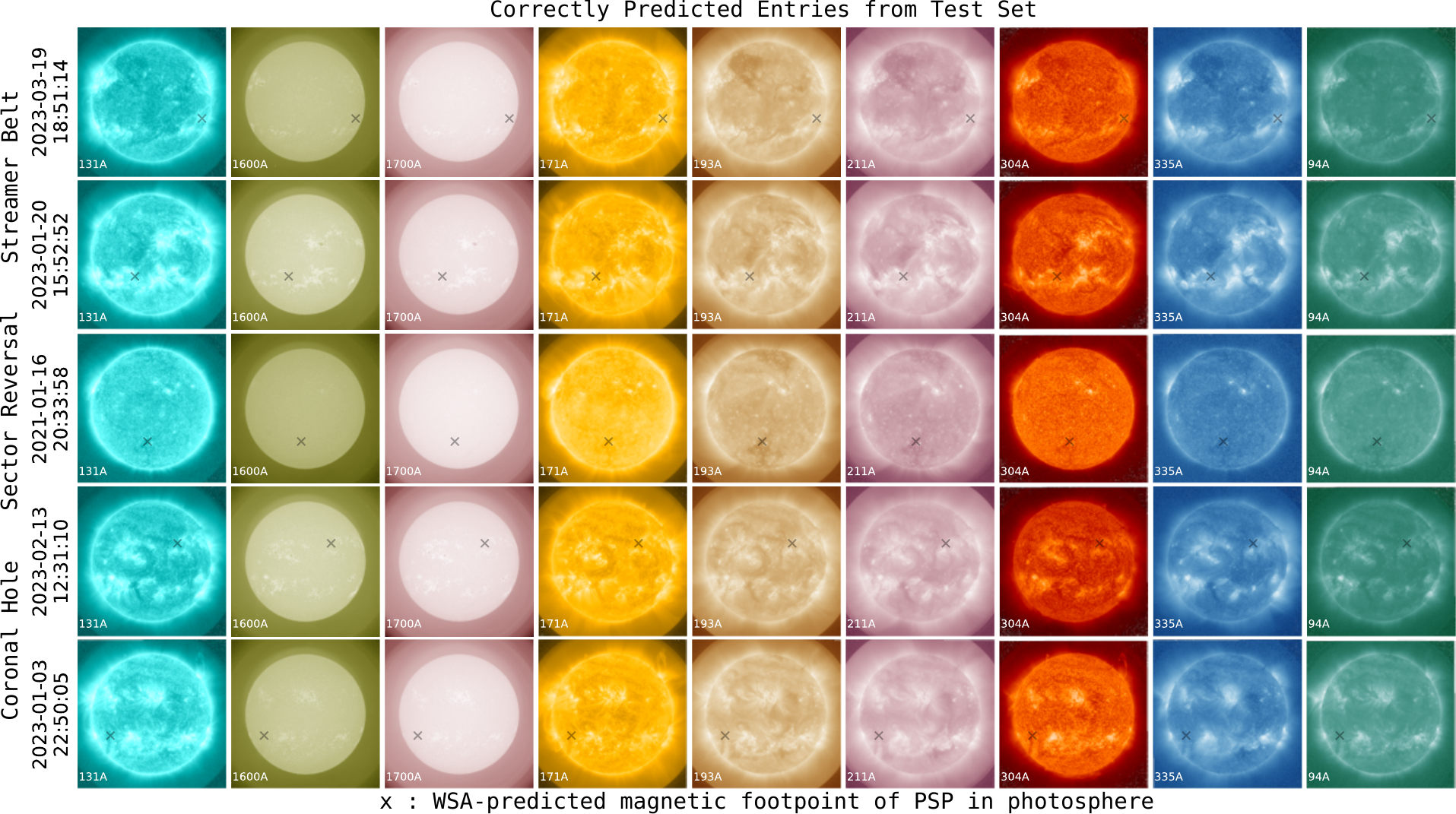}
    \caption{Examples of correct predictions on the test set for three solar wind classes: coronal hole, sector reversal, and streamer belt. Each column shows an SDOML/AIA image, with crosses indicating the predicted magnetic footpoints from PSP.}
    \label{fig:correctimages}
\end{figure*}

Figure~\ref{fig:correctimages} illustrates randomly selected correct predictions of the coronal hole, sector reversal, and streamer belt classes. For the coronal hole cases, the WSA-predicted magnetic footpoints lie in dark regions in the AIA images (especially visible in the 193~Å), which are coronal holes. For the streamer belt cases, the magnetic footpoints lie in the brighter, more dense regions of the Sun, characteristic of streamers. In the sector reversal case, the magnetic footpoints should lie around the area of the heliospheric current sheet.

\begin{table*}[h]
\centering
{%
\begin{tabular}{llcccc}
\hline
Head Type       & Training Strategy                              & Accuracy & Precision & Recall & F1-score    \\ \hline
                & Head Only (Frozen Backbone, Pretrained)        & 0.305    & 0.261     & 0.305  & 0.242 \\
Skip-connection & Backbone \& Head (Pretrained, Fine-tuned)      & 0.345    & 0.304     & 0.345  & 0.308 \\
                & Backbone \& Head (Random Initialization)       & 0.291    & 0.296     & 0.291  & 0.288 \\ \hline
                & Head Only (Frozen Backbone, Pretrained)        & 0.314    & 0.287     & 0.314  & 0.288 \\
Linear          & Backbone \& Head (Pretrained, Fine-tuned)      & 0.325    & 0.293     & 0.325  & 0.298 \\
                & Backbone \& Head (Random Initialization)       & 0.320    & 0.327     & 0.320  & 0.314 \\ \hline
\end{tabular}%
}
\caption{Test set performance of linear and skip-connection heads under three training strategies: frozen pretrained backbone, fine-tuned pretrained backbone, and random initialization. Metrics include Accuracy, Precision, Recall, and F1-score.}
\label{table:results}
\end{table*}

Table \ref{table:results} summarizes the performance of the evaluated architectures under three training regimes: training from scratch, transfer learning, and fine-tuning. The linear head achieves higher F1 when trained from scratch, while the skip-connection head performs best under fine-tuning. Overall, the linear head yields the highest accuracy across all settings. The model struggles to distinguish between streamer belt and sector reversal plasma classes, reflecting feature overlap and limitations of the \cite{xu_new_2015} labeling scheme \cite{camporeale2017classification}. Additionally, due to the scarcity of ejecta samples and WSA capabilities, the model does not learn a reliable representation for this class.

Although overall accuracy and F1 scores remain modest, the model appears to learn meaningful information from the AIA images. It seems to integrate features of the solar surface with the position of PSP (which is not in the field of view of the images) and the spacecraft's magnetic footpoints in the photosphere to inform its predictions. This suggests that even with limited performance, the pretrained embeddings capture relevant solar structures, highlighting the potential of foundation models for downstream in situ tasks.

\begin{figure*}[ht]
    \centering
    \includegraphics[width=0.80\linewidth]{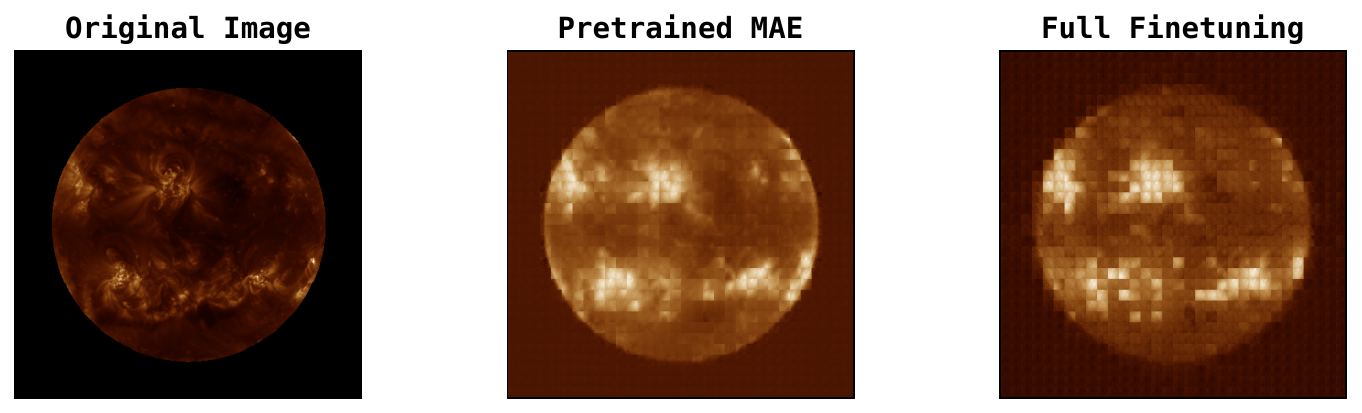}
    \caption{Example of image reconstruction using the MAE decoder.\textbf{ Left:} original SDOML/AIA 193~Å image (2023-01-03 05:38:00). \textbf{Middle:} reconstruction with the pretrained MAE without fine-tuning. \textbf{Right:} reconstruction after fine-tuning with a linear classification head.}
    \label{fig:image-comparison}
\end{figure*}

Figure \ref{fig:image-comparison} shows an example of one SDOML/AIA 193~Å image and its reconstructions by the pretrained MAE and the fully fine-tuned model. Fine-tuning enhanced contrast, making active regions brighter and coronal holes darker compared to the pretrained MAE output, highlighting the model's adaptation to task-specific features.

\subsection{Discussion and limitations}  
A primary factor limiting performance is the labeling scheme itself. The \cite{xu_new_2015} scheme provides a simple empirical segmentation, but its use of fixed thresholds introduces ambiguity near transitional regimes where plasma properties overlap, a common limitation of threshold-based classification methods in the continuously varying solar wind environment. This ambiguity is particularly evident between the streamer belt and sector reversal plasma classes, whose boundaries are not sharply defined, and is further compounded by the scarcity of ejecta samples, preventing the model from learning a robust representation for that class. As a result, overall classification accuracy remains modest ($\approx$30\%). The high similarity between the streamer belt and sector reversal classes suggests that some apparent misclassifications may not reflect a lack of learned structure, but rather the intrinsic ambiguity of the labels. The model exhibits a tendency to predict the streamer belt category, which is expected given its high prevalence in the dataset and the intrinsic similarity with the sector reversal class. While combining the high prevalence classes has not been explicitly tested, it indicates that the model could be capturing meaningful relationships between coronal features and in situ plasma measurements beyond merely predicting the majority class.

Another factor potentially limiting performance is the backbone itself. The SDO-FM model used here was pretrained exclusively on AIA images, capturing primarily coronal intensity features rather than magnetic topology. As a result, the model may have limited ability to fully represent the magnetic drivers of the solar wind. While this choice was dictated by the availability of pretrained foundation models, it highlights an avenue for future work where incorporating HMI magnetograms could improve downstream classification.

The mapping of PSP measurements to photospheric footpoints via the WSA model is inherently approximate, and the subsequent resampling of data from 8-hour to 1-minute cadence may introduce interpolation artifacts and noise. Also, as mentioned earlier, the WSA backmapping does not apply for the ejecta category. These approximations affect the fidelity of the positional encoding, which could in turn limit the discriminative power of the head architectures.

Overall, while this model is not intended as an operational or immediately deployable system, it provides a valuable proof of concept and diagnostic tool. Its limited performance helps reveal where current assumptions, such as coarse threshold-based labeling, fall short, offering insights that can guide future developments in both modeling and data annotation. In this sense, the present work contributes not only a methodology, but also a critical reflection on existing empirical schemes in heliophysics.

\section{Conclusion}
In this work, we adapted a foundation model for solar physics, combining MAE embeddings with neural field-based head architectures, to bridge remote-sensing imagery from the SDO with in situ plasma measurements from PSP for the classification of solar wind structures. Although the classification performance remains modest, the results demonstrate a proof of concept that positional encodings of spacecraft location and magnetic connectivity can enrich pretrained embeddings for downstream heliospheric tasks.

This study demonstrates that foundation models can serve as a bridge between solar imagery and heliospheric in situ observations, an essential and nontrivial step toward unified models for solar physics and space weather forecasting. The presented pipeline represents a first stone in exploring how foundation model embeddings can be extended from solar remote sensing to in situ applications. While current results are limited, they reveal that embeddings learned from coronal imagery capture meaningful physical information that can inform solar wind classification. This connection between modalities highlights a promising pathway toward more robust, interpretable, and data-driven approaches to space weather analysis.

\section{Future Work}  
Several avenues could be explored to address the limitations identified in this study. First, improving the labeling scheme by incorporating hybrid or probabilistic methods could reduce ambiguity near transitional regimes and better capture the diversity of solar wind structures. Second, extending the foundation model pretraining to include additional magnetic field data, such as HMI magnetograms, could enhance the transferability of embeddings for in situ tasks. Third, larger and more balanced test and validation splits, along with metrics tailored for class imbalance, would provide a more robust assessment of model performance. Finally, optimizing data storage and loading pipelines could improve training scalability for larger datasets and more complex models, enabling more comprehensive experimentation.

\section*{Acknowledgements}
This work is a research product of Heliolab (heliolab.ai), an initiative of the Frontier Development Lab (FDL.ai). FDL is a public–private partnership between NASA, Trillium Technologies (trillium.tech), and commercial AI partners including Google Cloud and NVIDIA. Heliolab was designed, delivered, and managed by Trillium Technologies Inc., a research and development company focused on intelligent systems and collaborative communities for Heliophysics, planetary stewardship and space exploration. We gratefully acknowledge Google Cloud for extensive computational resources, and NVIDIA Corporation for access to DGX Cloud, enabled through NVIDIA and VMware. This material is based upon work supported by NASA under award No. 80GSFC23CA040. Any opinions, findings, and conclusions or recommendations expressed are those of the author(s) and do not necessarily reflect the views of the National Aeronautics and Space Administration.

\bibliographystyle{unsrt}  
\bibliography{references}

\end{document}